\documentclass[letterpaper]{article} % DO NOT CHANGE THIS
\usepackage{aaai22}  % DO NOT CHANGE THIS
\usepackage{times}  % DO NOT CHANGE THIS
\usepackage{helvet}  % DO NOT CHANGE THIS
\usepackage{courier}  % DO NOT CHANGE THIS
\usepackage[hyphens]{url}  % DO NOT CHANGE THIS
\usepackage{graphicx} % DO NOT CHANGE THIS
\usepackage{natbib}  % DO NOT CHANGE THIS AND DO NOT ADD ANY OPTIONS TO IT
\usepackage{caption} % DO NOT CHANGE THIS AND DO NOT ADD ANY OPTIONS TO IT
\DeclareCaptionStyle{ruled}{labelfont=normalfont,labelsep=colon,strut=off} % DO NOT CHANGE THIS
\usepackage{soul}
\usepackage{xcolor}

\usepackage{algorithm}
\usepackage{algorithmic}

\usepackage{newfloat}
\usepackage{listings}
\floatstyle{ruled}
\newfloat{listing}{tb}{lst}{}
\floatname{listing}{Listing}
\usepackage{graphicx} 
\graphicspath{{images/}}
\usepackage{caption}
\usepackage{subcaption}
\usepackage{amsmath,amsthm,amssymb}

\title{Automatic Structured Pruning for Efficient Architecture in Federated Learning}
\author{
   Thai Vu Nguyen, Long Bao Le, Anderson Avila \textsuperscript{\rm 1}
}
\affiliations{
   \textsuperscript{\rm 1}INRS-EMT, University of Quebec\\
}

\usepackage{bibentry}
\begin{document}

\maketitle

\begin{abstract}
% Federated Learning (FL) is a distributed learning paradigm that enables multiple clients to collaboratively train deep learning models while preserving data privacy.
In Federated Learning (FL), training is conducted on client devices, typically with limited computational resources and storage capacity.
To address these constraints, we propose an automatic pruning scheme tailored for FL systems.
Our solution improves computation efficiency on client devices, while minimizing communication costs. 
One of the challenges of tuning pruning hyper-parameters in FL systems is the restricted access to local data. 
Thus, we introduce an automatic pruning paradigm that dynamically determines pruning boundaries. 
% To address this constraint, we propose an automatic pruning scheme for FL systems, enhancing efficiency in both training on client devices and communication costs on the FL system. 
% Given the restricted access to local data in FL, which poses significant challenges in tuning pruning hyper-parameters such as sparsity level, 
Additionally, we utilized a structured pruning algorithm optimized for mobile devices that lack hardware support for sparse computations.  Experimental results demonstrate the effectiveness of our approach, achieving accuracy comparable to existing methods. Our method notably reduces the number of parameters by 89\% and FLOPS by 90\%, with minimal impact on the accuracy of the FEMNIST and CelebFaces datasets. 
Furthermore, our pruning method decreases communication overhead by up to 5x and halves inference time when deployed on Android devices.
\footnote{The code implementation is available online: \url{https://github.com/NguyenThaiVu/prune_fl_project}.}
\end{abstract}

\section{Introduction}

Recently, we have witnessed an increasing interest in shifting cloud computing to edge computing \cite{li2021talk}. 
In fact, bringing computation to the edge of computer networks can reduce latency, benefiting real-time applications, such as autonomous driving \cite{jin2024fractional}. 
On the other hand, the data generated, at an unprecedented rate, from billions of edge devices can be used for training and improving AI models \cite{jia2024fedlps,nguyen2023attention}. 
In such scenarios, Federated Learning (FL) has emerged as a promising alternative to process such an extensive amount of data while preserving users' privacy \cite{mcmahan2017communication}. 
This is achieved by decentralizing the training process of machine learning (ML) models and keeping personal user data on clients' devices. 
Thus, training is performed locally and only the model parameters are sent to the central server. The final global model is attained via aggregation of the parameters from all clients. 
% In the traditional centralized learning process, training takes place typically in a powerful central server with hundreds of GPUs.

In the FL settings, where the primary training process occurs on devices with limited computational and storage resources, there is growing interest in developing strategies to decrease the model's footprint.
Several studies, for instance, have proposed neural network pruning to reduce model size. 
The frameworks proposed in \cite{li2021lotteryfl} and \cite{isik2023sparse} apply the lottery ticket hypothesis to the FL settings. 
The work presented in \cite{qiu2022zerofl} selects the top-k weights of local models to be sent back to the central server for aggregation.
In \cite{Bibikar_Vikalo_Wang_Chen_2022}, authors propose a sparse training approach tailored specifically for the FL settings.  
A crucial limitation of such approaches is that they require pruning to be performed on the local devices. 
This results in higher computational costs when compared to the normal local training.  
Additionally, the differences in local models lead to different pruning masks, complicating the aggregation of these masks into the unified global model. 
To overcome these problems, we propose a pruning procedure that is executed on the central server rather than on individual local devices.

It is important to note that the aforementioned works rely on unstructured pruning, which removes individual weight elements and results in a sparse network with many zero values. 
Performing fast computations on sparse matrices requires support from specialized libraries, such as cuSPARSE, or hardware (e.g., NVIDIA Ampere GPU) and clients' devices participating in FL training often lack such requirements. 
Additionally, maintaining sparse data structures requires extra storage for information like compressed sparse rows or binary masks. 
Thus, inspired by the structured pruning technique presented in \cite{li2017pruning}, which prunes entire filters rather than individual weights, this paper proposes pruning filters in convolutional networks, maintaining the dense structure of the global model while ensuring compatibility with the simple computational capabilities of client devices.

% Another limitation of the aforementioned works is that they use unstructured pruning, which removes individual weight elements and results in a sparse network with many zero values. 
% Performing fast computations on sparse matrices requires support from specialized libraries (cuSPARSE) or hardware (NVIDIA Ampere GPU). 
% However, in the FL system, client devices are often simple IoT or mobile devices lacking these requirements. 
% Additionally, maintaining sparse data structures requires extra storage for information like compressed sparse rows or binary masks.
% This paper is inspired by the structured pruning technique \cite{li2017pruning}, which prunes entire filters rather than individual weights. 
% By pruning filters in the convolutional network, we maintain the dense structure of the global model, ensuring compatibility with the simple computational capabilities of client devices.

\begin{figure*}[htp]
    \centering
    \includegraphics[width=0.85\textwidth]{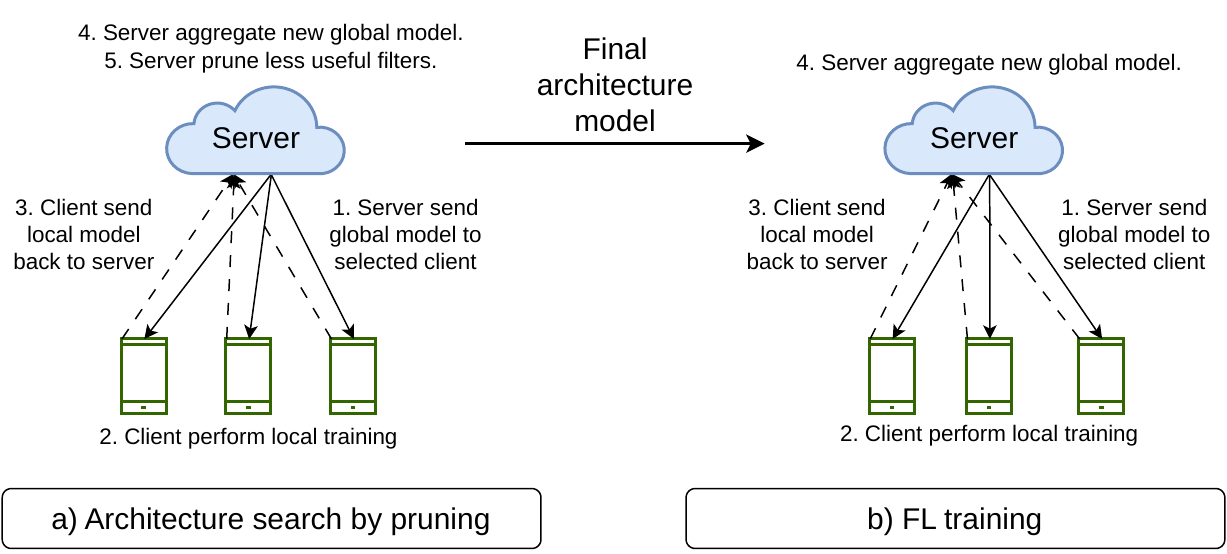}
    \caption{An overview of our pruning scheme in Federated Learning system.}
    \label{fig:overall_flow}
\end{figure*}

While centralized learning allows access to the training data and therefore proper tuning of pruning hyper-parameters, such as the sparsity fraction and the number of pruned layers, decentralized learning presents the inability to examine the whole training data directly during training. 
This makes it difficult to choose a suitable model architecture. 
For example, in the FL settings, we cannot perform a grid search to determine the optimal number of filters in a convolutional neural network (CNN). 
To address this challenge, we propose an automatic pruning algorithm, which dynamically defines the pruning boundaries and removes redundant filters without disrupting the FL procedure. 

The proposed framework for pruning CNNs in the FL settings is illustrated in Figure~\ref{fig:overall_flow}. 
The FL training procedure is divided into two sub-stages. 
First, we perform automatic pruning to identify the well-suited architecture, as depicted in Figure~\ref{fig:overall_flow}-a.
This phase involves dynamically eliminating unnecessary filters to reduce the model's size without yet considering the accuracy of the global model.
After obtaining the final global model architecture, we proceed with the standard FL training phase (Figure~\ref{fig:overall_flow}-b).
The second stage is dedicated to training the final global model to achieve the desired accuracy for the specific task.

Below, we summarize the four main contributions of our work:
\begin{itemize}
    \item We designed a pruning scheme for FL that significantly reduces redundant parameters/FLOPS in the global model while maintaining the desired accuracy. Additionally, the pruning procedure runs on the central server, reducing computational burdens for local clients.

    \item We proposed a structured pruning algorithm that automatically determines the pruning boundary and preserves a dense network structure, ensuring fast computation on basic client hardware.
    % \item We proposed an algorithm that automatically determines the pruning boundary, addressing the overparameterization issue in the model. In addition, we employ a structured pruning approach to preserve a dense network, ensuring fast computation on simple client hardware.
    
    \item Extensive experiments show that our pruning algorithm is effective across various architectures, including CNNs, ResNet, and Inception. Additionally, the pruned models significantly reduce inference time on real-world Android devices.
    
    \item We also present supplemental studies showing improvements in communication costs and consistency when training with varying random numbers of selected clients in the FL system.
\end{itemize}

The rest of the paper is organized as follows. Section \ref{sec_background} presents the related work in network pruning, besides  summarizing the differences between our paper and existing work.
Section \ref{sec_prune_fl} introduces our proposed method to prune convolutional neural networks in the FL settings.
In Section \ref{sec_experiment}, we conduct extensive experiments to validate the superior performance of our method, followed by the concluding remarks in Section~\ref{sec_conclusion}.

\section{Related Works}
\label{sec_background}

\subsection{Network Pruning in Federated Learning}

Pruning techniques are widely used to reduce the size of neural network models. These approaches are crucial to facilitate the execution of such models on resource-constrained devices (e.g., mobile and IoT gadgets).
Pruning methods can be categorized into two main types: unstructured pruning \cite{han2015learning} and structured pruning \cite{li2017pruning}.
Unstructured pruning involves removing individual weights in the network, resulting in a sparse weight matrix. 
In contrast, structured pruning implies the elimination of entire components of the network, such as layers or filters, while maintaining a dense network architecture.

% There have been significant works using unstructured pruning to remove individual weights within the neural network of the FL system. 
Due to the decentralized nature of FL, applying unstructured pruning on each client may result in differently pruned networks, leading to inconsistencies during the aggregation of global models.
Some studies have leveraged these varying pruning masks for personalized federated learning \cite{li2021fedmask,li2021lotteryfl,dai2022dispfl}. 
Other researchers \cite{Bibikar_Vikalo_Wang_Chen_2022,isik2023sparse,qiu2022zerofl,babakniya2023revisiting,jiang2023computation,jiang2023complement,huang_distributed_2023} have proposed methods to aggregate different pruning masks into the global sparse model. 
However, the resulting global model still contains a sparse weight matrix, which can lead to slower inference times on limited client devices.
% In \cite{huang_distributed_2023}, the proposed method appears to offer no saving in local storage and computation. Because it requires sending C candidate models to each clients and performing computations on the clients.

On the other hand, structured pruning emerges as an ideal choice for deployment on edge devices in the FL system.
This method preserves dense network architectures, facilitating straightforward deployment on simple hardware without the need for specialized sparse matrix support.
In \cite{vahidian2021personalized,wu2020fedscr}, for instance, the pruning process occurs at the edge of the network, increasing the computational burdens of less powerful mobile/IoT devices. 
In contrast, the authors in \cite{xu2021accelerating} propose to perform pruning on a central server. Their method prunes the model only once, during the initialization of the global model, excluding the pruning procedure from the training process.
The method presented in \cite{zhang2022fedduap} assumes that the training dataset is located on the central server, which contradicts the decentralized nature of FL
The FedPara method \cite{hyeon2021fedpara} employs the low-rank Hadamard product to reduce the number of parameters. Nevertheless, the low-rank matrix approximation still occurs on client devices, increasing local computation.
% Based on these observations, we propose that the pruning procedure should be conducted on the central server. 
% This approach would alleviate the computational load on client devices while maintaining the benefits of structured pruning.

Unlike previous work, our method performs structured pruning on a powerful central server, offering several advantages.
First, structured pruning maintains the dense network structure, facilitating straightforward hardware implementation. 
Second, by conducting the pruning procedure on the central server, we eliminate additional computational burdens on client devices. 
Finally, because pruning is performed centrally, our method does not require an aggregation mechanism to combine different pruned client models into a global model. 
Instead, we utilize the standard FedAvg algorithm for model aggregation.

\subsection{Pruning Hyper-parameter}
\label{prune_architec_fl}
% a. Pruning acts as the architecture search during FL.
% b. Automatic pruning, rather than predefined threshold.

Previous works primarily rely on the pre-defined pruning hyper-parameters such as accuracy threshold, sparsity ratio, or pruning rate to control the pruning procedure.
The method \cite{li2021lotteryfl} employs a predefined pruning rate to determine the number of pruned parameters per client. However, selecting an appropriate pruning rate is challenging, as a rate effective for current devices may not be suitable for others.
PruneFL \cite{jiang_model_2023} relies on powerful and trusted clients to find suitable architecture. However, finding powerful clients is challenging in a heterogeneous FL system.
Other methods \cite{Bibikar_Vikalo_Wang_Chen_2022,qiu2022zerofl,babakniya2023revisiting} utilize pre-defined sparsity density, or sparsity ratio to manage the pruning procedure. 
The FedDUAP \cite{zhang2022fedduap} adopts server data to halt the pruning method. 
Additionally, methods \cite{li2021fedmask,hyeon2021fedpara} rely on pre-defined lower rank values or hard thresholds to reduce the number of parameters. 
However, in the decentralized FL, selecting appropriate hyper-parameters poses a challenge due to the absence of central training data, which are critical for determining optimal hyper-parameters.

Unlike previous approaches, our automatic pruning algorithm simplifies the hyper-parameter selection process by removing the need to specify sparsity ratios or pruning rates. 
Instead, it automatically identifies and removes redundant filters within the model.
Additionally, by not assigning specific sparsity ratios to each client, we prevent pruning discrepancies that can occur when a pre-defined ratio is suitable for one client but inappropriate for others.
This approach ensures consistent performance of our pruning algorithm across heterogeneous clients.

% Our works proposed the pruning method (Algorithm \ref{algo:prune}), which automatically performs the pruning process without hyper-parameters such as sparsity ratio, or pruning rate.
% Based on insights from \cite{liu2018rethinking}, the value of structured pruning lies in identifying an efficient architecture.
% From that observation, we have restructured the FL training procedure into two stages: architecture search and FL training. The primary objective of the first stage is to iteratively prune the model until the most suitable architecture is determined. Secondly, once the final architecture is identified, we transition to the regular FL training.

\section{Structured Pruning ConvNets for Federated Learning}
\label{sec_prune_fl}

% In this section, we will describe the algorithm for automatic pruning filter (subsection \ref{prune_crite}) and outline the procedure for pruning network architecture in FL training (subsection \ref{prune_search}). Moreover, subsection \ref{sec_analysis} validates that the remained filters still retain statistical properties of good weight initialization.  

In this section, we present the proposed two stages FL training procedure. We first describe the automatic pruning filter algorithm used in the first stage. We then outline the procedure applied in the second stage for pruning network architecture in FL training. We end by showing that the remained filters still retain statistical properties of good weight initialization.

\subsection{Automatic Pruning Algorithm}
\label{prune_crite}

There are two approaches for determining the number of layers to prune: uniform pruning and automatic pruning \cite{liu2018rethinking}.
In uniform pruning, the same number of weights/filters is removed at each epoch throughout the training process. 
The uniform pruning works well in centralized learning, where the entire training dataset is fully accessible, facilitating the selection of optimal architecture based on validation performance criteria.
However, in the context of FL, automatic pruning becomes an intelligent strategy due to the challenge of selecting optimal pruning hyper-parameters, without entirely access to the training data.

% This approach necessitates the predetermined quantity of weights or filters to be pruned.
% For example, the work by \cite{li2021lotteryfl} uses the fixed pruning rate $r_p$, \cite{qiu2022zerofl} retains only the top-k weights, and \cite{Bibikar_Vikalo_Wang_Chen_2022} maintains the pre-defined sparsity level $S$.
% In contrast, automatic pruning systematically identifies and removes redundant filters from the model without requiring predefined pruning hyper-parameters.

\begin{figure}[htp]
    \centering
    \includegraphics[width=0.45\textwidth]{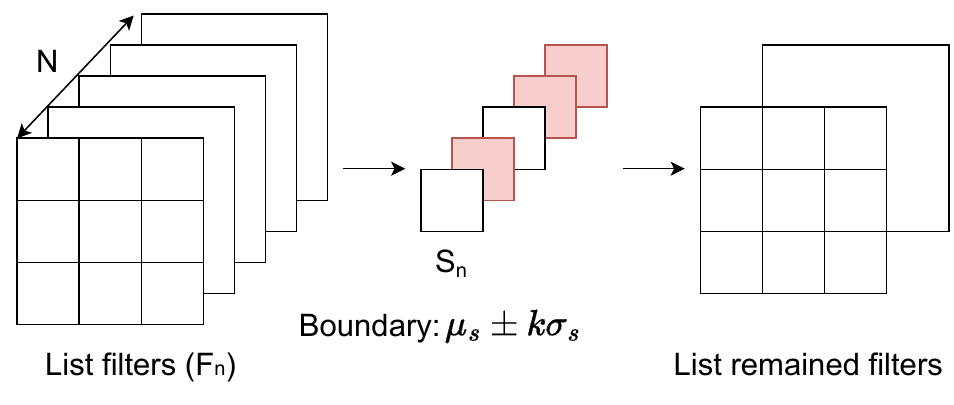}
    \caption{An overview of our structured pruning algorithm, which assesses convolution filters based on the sum of their absolute values $S_n$. Filters that fall outside the boundary $\mu_s \pm k \sigma_s$ (where $\mu_s$ is the mean and $\sigma_s$ is the standard deviation of $S_n$) are pruned and highlighted in red.}
    \label{fig:prune_filter}
\end{figure}

To effectively perform automatic pruning, we proposed Algorithm~\ref{algo:prune}. 
This algorithm not only automatically eliminates redundant filters, but also retains the dense structure of the model architecture.
The illustration of the pruning algorithm is described in Figure~\ref{fig:prune_filter}. 
Although Algorithm \ref{algo:prune} is specifically  described for 2D convolutional filters, it can be extended to other types of convolutional filters, such as 1D and 3D filters. 
For 1D filters, the algorithm would sum the absolute values of elements along a single dimension. 
For 3D filters, the algorithm would sum the absolute values across three dimensions.
In both cases, the computation of the mean, standard deviation, and pruning criteria should be appropriately adapted to account for the dimension of the filters.

\begin{algorithm}[ht]
    \caption{Pruning filter Algorithm}
    \label{algo:prune}

    \textbf{Input}: 
    \begin{itemize}
    \item  $W \in \mathbb{R}^{N \times K \times K}$: initialized weight matrix  
    \item  $N$: number of 2D filters
    \item  $K \times K$: filter dimension
    \item  $k$: constant defining pruning boundary  
    \end{itemize}
    
    \textbf{Output}: 
    \begin{itemize}
        \item $W' \in \mathbb{R}^{N' \times K \times K}$: remained weight matrix
    \end{itemize}

    \textbf{Algorithm}: 
    \begin{algorithmic}[1]
        \STATE Initialize $S$ as a vector of length $N$
        \FOR{$n = 1$ to $N$}
            \STATE $S_n = \sum_{i=1}^{K} \sum_{j=1}^{K} |W_{nij}|$
        \ENDFOR

        \STATE Mean $\mu_s = \frac{1}{N} \sum_{n=1}^{N} S_n$ 
        \STATE Standard deviation $\sigma_s = \sqrt{\frac{1}{N} \sum_{n=1}^{N} (S_n - \mu_s)^2}$ 
        
        \STATE Lower bound $= \mu_s - k \sigma_s$ 
        \STATE Upper bound $= \mu_s + k \sigma_s$ 

        \STATE Initialize an empty list $W'$ to store remained filters
        \FOR{$n=1$ to $N$}
            \IF{$\mu_s - k \sigma_s \leq S_n \leq \mu_s + k \sigma_s$}
            \STATE Add $W_n$ to $W'$
        \ENDIF
        \ENDFOR
        
        \STATE Return $W'$
    \end{algorithmic}
\end{algorithm}

The utilization of the mean ($\mu_s$) and standard deviation ($\sigma_s$) of the sum absolute values ($S_n$) across all filters to establish a pruning boundary offers a statistical intuition for assessing the influence of filters on the prediction. 
Filters exhibiting lower $S_n$ values contribute minimally to the output, suggesting limited contribution. Conversely, exceptionally high $S_n$ values may introduce excessive noise, potentially disrupting the output. 
By incorporating a scaling factor $k$ into the boundary definition, the algorithm allows for flexible adjustment between aggressive pruning (lower $k$) and performance retention (higher $k$).

\begin{figure}[htp]
    \centering
    \includegraphics[width=0.4\textwidth]{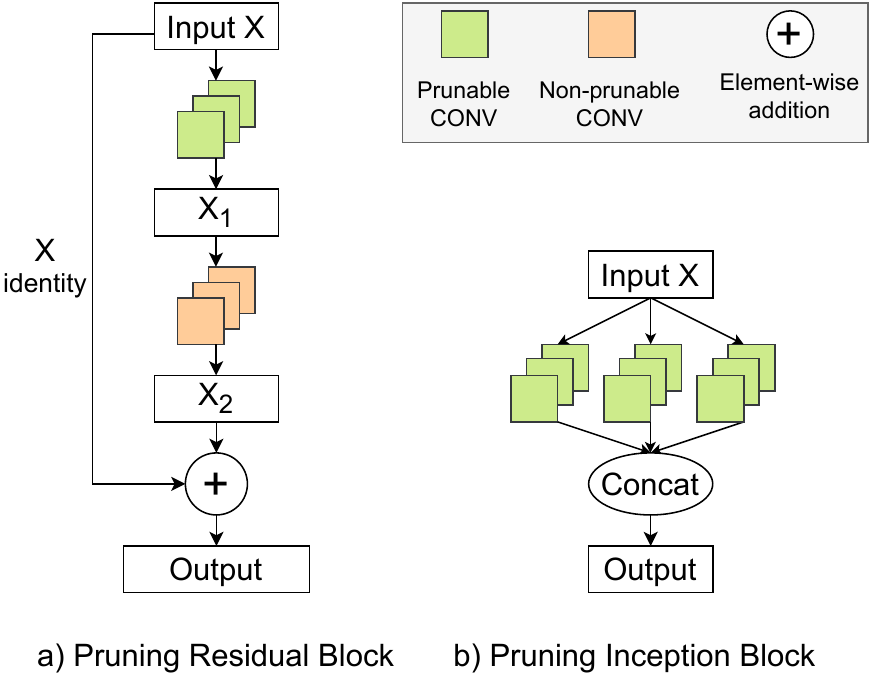}
    \caption{An overview of pruning procedure for complex designed convolutional architecture. Prunable Conv indicates layers can be pruned by Algorithm \ref{algo:prune}. Non-prunable Conv is layers, which have a fixed number of filters and can not be pruned.}
    \label{fig:prune_complex_filter}
\end{figure}

For straightforward CNN architectures (AlexNet or VGG network), layer-by-layer pruning is easily applicable, as discussed in Algorithm \ref{algo:prune}. 
However, in more complicated architectures, such as residual blocks \cite{he2016deep} or inception modules \cite{szegedy2015going}, pruning filters become more complex. 
Firstly, in a residual block (Figure~\ref{fig:prune_complex_filter}.a), pruning the first convolutional layer is permissible because it does not alter the output shape. 
However, pruning the second convolutional layer is not advisable, as it directly affects the output shape of the residual block, potentially disrupting the entire network.
Secondly, in inception modules (Figure~\ref{fig:prune_complex_filter}.b), all convolutional layers can be pruned independently. This flexibility stems from the inception module's design, which employs concatenation operations at the end rather than requiring consistent shapes through element-wise operations.
As a result, pruning filters within an inception block can be done individually without affecting other parts of the network.

\subsection{Decentralized Training Procedure}
\label{prune_search}

The overall procedure of the proposed pruning method in the FL system is illustrated in Figure~\ref{fig:overall_flow}, which includes two main stages: architecture search by pruning and FL training. 

\textbf{Stage 1: Architecture search by pruning.}
% As discussed in subsection \ref{prune_architec_fl}, the main purpose of pruning is to identify the most suitable architecture. 
The automatic pruning Algorithm \ref{algo:prune} is applied to the global model at each round of FL training.
The detailed procedure for the architecture search stage is depicted in Figure~\ref{fig:overall_flow}-a, and is described as follows:
\begin{enumerate}
    \item The server sends a global model to the selected subset of clients. 
    \item Clients receive the global model and perform local training on private datasets. 
    \item All clients send back the trained local model to the server.
    \item The server aggregates all local models into the new global model.
    \item The server prunes the new global model using the automatic pruning Algorithm \ref{algo:prune}.
\end{enumerate}

% All client have same model architecture => Do not need mechanism to aggregation.
% Prune at server.
% Stop prune.

\begin{table*}[t]
    \centering
    \begin{subtable}{\textwidth}
        \centering
        \caption{Performance on FEMNIST dataset}
        \begin{tabular}{lc|cc|cc|c}
            \hline
            Architecture & \# Conv & Number & Pruned \%  & FLOPS  & Pruned \% &  Accuracy \\
            & &  params &  params &   & FLOPS & \\
            \hline
            Conv & 2 & 56 126 &  & $7.5 \times 10^6$ &  & 74.72 \\
            Conv-pruned & 2 & 16 683 & 70\% & $2.2 \times 10^6$ & 70\% & 74.22 \\
            \hline
            ResNet & 7 & 33 214 & & $5.2 \times 10^5$ & & 79.29 \\
            ResNet-pruned & 7 & 5 668 & 82\% & $0.9 \times 10^5$ & 82\% & 74.55 \\
            \hline
            Inception & 9 & 246 236 & & $31 \times 10^6$ & & 80.19 \\
            Inception-pruned & 9 & 26 744 & 89\% & $3 \times 10^6$ & 90\% & 78.81 \\
            \hline
        \end{tabular}
    \end{subtable}\\
    
    \begin{subtable}{\textwidth}
        \centering
        \caption{Performance on CelebFaces dataset.}
        \begin{tabular}{lc|cc|cc|c}
            \hline
            Architecture & \# Conv & Number & Pruned \%  & FLOPS  & Pruned \% &  Accuracy \\
            & &  params &  params &   & FLOPS & \\
            \hline
            Conv & 4 & 80 930 & & $152 \times 10^6$ & & 95.87 \\
            Conv-pruned  & 4 & 10 998 & 86\% & $29 \times 10^6$ & 80\% & 92.59 \\
            \hline
            ResNet & 7 & 30 626 & & $5.2 \times 10^6$ & & 89.50 \\
            ResNet-pruned & 7 & 6 791 & 77\% & $1.6 \times 10^6$ & 69\% & 85.83 \\
            \hline
            Inception & 9 & 243 752 & & $324 \times 10^6$ & & 87.78 \\
            Inception-pruned & 9 & 127 080 & 47\% & $171 \times 10^6$ & 47\% & 86.23  \\
            \hline
        \end{tabular}
    \end{subtable}
        
    \caption{The results of our pruning method on the FEMNIST and CelebFaces datasets across Convolution, ResNet, and Inception architecture. We report the highest test accuracy achieved.}
    \label{tab:main}
\end{table*}

In the above process, steps 1-5 are repeated until discovering a suitable architecture. 
We adopt a simple early stopping schedule to stop the architecture search stage: after $c$ rounds ($c$=3 in the experiment), if the number of parameters in the global model does not decrease, we halt the architecture search stage. 
That showcases another advantage of automatic pruning: there is no need to explicitly specify the number of pruning rounds, as the determination is made automatically during the training process.

In step 4, we utilize the standard FedAvg \cite{mcmahan2017communication} on each round to aggregate all trained local models into a new global model. 
Since clients do not prune their local models, all of which share the same architecture, no additional aggregation scheme is necessary.
Moreover, because pruning occurs on the server side, it mitigates potential issues for less powerful client devices. 

\textbf{Stage 2: Federated Learning training}.
After the first architecture search stage, we obtain the final global model architecture. 
Subsequently, we proceed with the standard Federated Learning training stage utilizing the selected architecture.
The primary objective of this second stage is to fully optimize the model's performance for the specific task we are working on.

\section{Experiments}
\label{sec_experiment}

\subsection{Experimental Setup}
\label{exp_setup}

We evaluate the performance of our pruning method on three convolutional neural network architectures: Vanilla Convolution, ResNet, and Inception. 
The filter size is 5x5, and the number of filters for each model is detailed in Table 1. 
All models were trained from scratch, using cross entropy as loss function and Adam as optimizer. 
To simulate real-world conditions as closely as possible, we use the standard LEAF benchmark \cite{caldas2018leaf} with the FEMNIST and CelebFaces datasets. The FEMNIST dataset includes more than 800,000 images with size of 28x28 pixels and the task is a digit classification problem with 62 output categories. The CelebFaces dataset includes RGB images with 82x82x3 pixels, with each sample containing a human face to be predicted  as male or female by the proposed model. On client devices, we limit the local training to 5 epochs with batch size equal 32 due to the constrained computing power of clients, particularly mobile and IoT devices.  Regarding the FL system, we randomly choose 10\% clients in each round, with a total of 500 rounds considered in our experiments.

\begin{figure*}[!htp]
    \centering
    \includegraphics[width=1\textwidth]{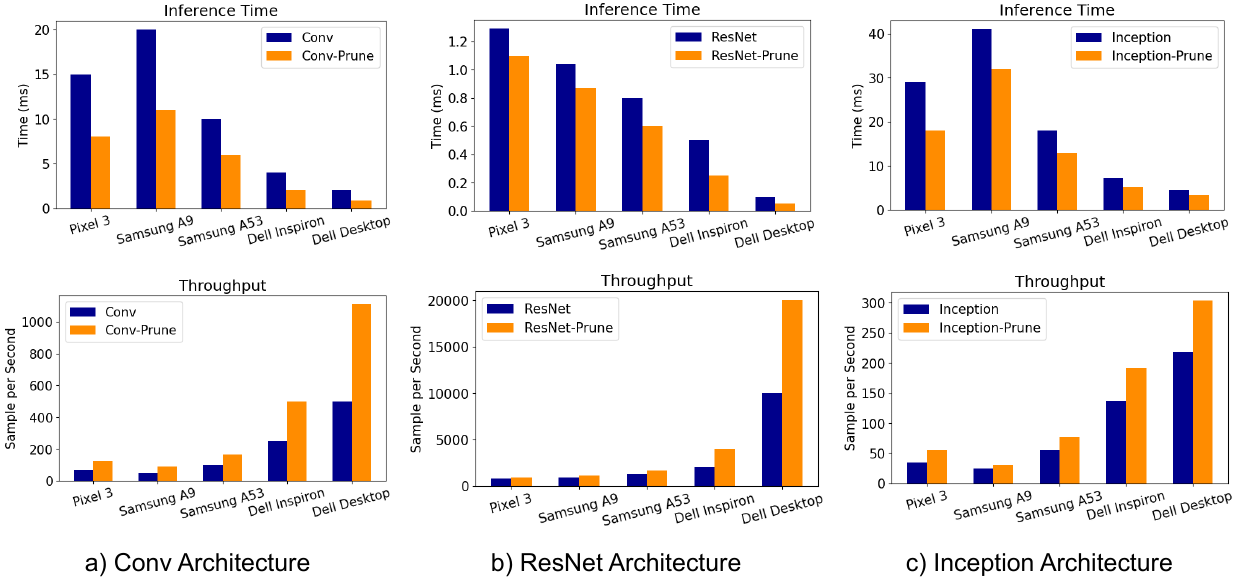}
    \caption{Evaluation of inference time (milliseconds) and throughput (samples per second) across five devices for various model architectures.}
    \label{fig:read_device}
\end{figure*}

\subsection{Baseline Results}
\label{exp_base}

We will begin by examining the impact of the pruned model in the Federated Learning (FL) system.  The primary objective of this comparison is to highlight the ability of our method to achieve maintain acceptable accuracy with minimal storage cost (number of parameter) computation cost (FLOPS). Table~\ref{tab:main} describes the number of parameters and FLOPS reduced after applying pruning. Results are attained from the FEMNIST and CelebFaces datasets. In all cases, our method significantly reduced the number of parameters/FLOPS and the accuracy is maintained to acceptable levels. For instance, in the case of the Inception architecture on the FEMNIST dataset, the number of parameters and FLOPS are reduced by 90\% while the accuracy is reduced by only 2\%. Despite the complexity of the CelebFaces dataset (200,000 RGB images), our pruning method still achieves promising results.
Notably, with the Conv architecture, the number of parameters is reduced by 86\% but the accuracy is only reduced by about 3\%.

\begin{table}
\centering
\caption{The performance comparison between our model and other pruning methods on the FEMNIST dataset.}
\begin{tabular}{c|c|c}
    \hline
    Method & Prune Type  &  Accuracy \\
    \hline
    LotteryFL & unstructured & 58.69 \\
    FedPM  & unstructured   & 61.29 \\
    FedPara & structured  & 65.52 \\
    PruneFL & unstructured & 57.09 \\
    FlashFL & unstructured & 65.39 \\
    % Sparse  & unstructured & \\
    FeDST & unstructured & 52.34 \\
    FeDST+FedProx  & unstructured & 52.94 \\
    Our model & structured & \textbf{74.22} \\
    \hline
    % Non-prune & - & 56 126 & $7.5 \times 10^6$ & 74.72 \\
\end{tabular}
\label{tab:baseline_compare}
\end{table}

In addition, we also compared the accuracy of our pruned model with other methods on the FEMNIST dataset, as shown in Table~\ref{tab:baseline_compare}. To ensure comparison closely follows the original papers, we used their publicly repository. The results show that our method achieves 74.22\% accuracy, which outperform other methods.
One explanation for this result is that our pruning method preserves the dense structure of the model, which facilitates the global model aggregation on the FL system.

\subsection{Deployment on Real Devices}
\label{exp_real_devices}

\begin{table}
\centering
\caption{The hardware information of several testing devices.}
\begin{tabular}{c|ccc}
    \hline
    Devices & Chip & CPU cores & RAM \\
    \hline
    Google Pixel 3 & Snapdragon 845 & 8 & 8 GB \\
    Samsung A9 & Snapdragon 660 & 8 & 6 GB \\
    Samsung A53 & Exynos 1280 & 8 & 6 GB \\
    Dell Inspiron 15 & Intel i7-7700 & 4 & 8 GB \\
    Dell Desktop & Intel i7-4790 & 8 & 16 GB \\
    \hline
\end{tabular}
\label{tab:real_device}
\end{table}

In this section, we present the results of deploying the pruned model on various real-world devices, such as smartphones, personal laptops, and desktops. On mobile scenarios, the pruned model was integrated into a Java-based mobile application for Android devices such as the Google Pixel 3, Samsung Galaxy A9, and Samsung Galaxy A53.
On desktop scenarios, the model was incorporated into a Python-based application using Streamlit, a framework designed for building interactive web applications, which was executed on a Dell Inspiron laptop and a Dell desktop. The reader can refer to Table~\ref{tab:real_device} for detailed information regarding the devices used herein. By embedding the pruned model into these devices, we were able to assess its performance in a simulated practical real-world environment. 

Android devices currently offer limited support for well-known deep learning frameworks such as TensorFlow and PyTorch. To handle this limitation, we converted the trained model into the TensorFlow Lite (tflite) format, which is a lightweight model format for mobile environments. For practical application, we utilize a model trained on the CelebFaces dataset to classify images as male or female. We evaluated the effectiveness of the pruning method by examining inference time and throughput.
Inference time refers to the duration required to perform a single inference operation on an image, excluding auxiliary tasks such as image loading and data processing. 
Throughput is defined as the number of samples (images) the model can process per second. This metric is crucial for evaluating the model’s ability to manage a high volume of predictions within a specific time frame.

Figure~\ref{fig:read_device} describes the comparison in terms of inference time and throughput between the original and the pruned models across four different devices. The results indicate that the pruned model cuts the inference time by nearly half and doubles the throughput on all tested devices, demonstrating a substantial improvement in efficiency. 
For instance, in a standard convolutional architecture, the throughput on a Dell Desktop increased from 500 to 1000 samples per second.

\begin{figure}[htp]
    \centering
    \includegraphics[width=0.4\textwidth]{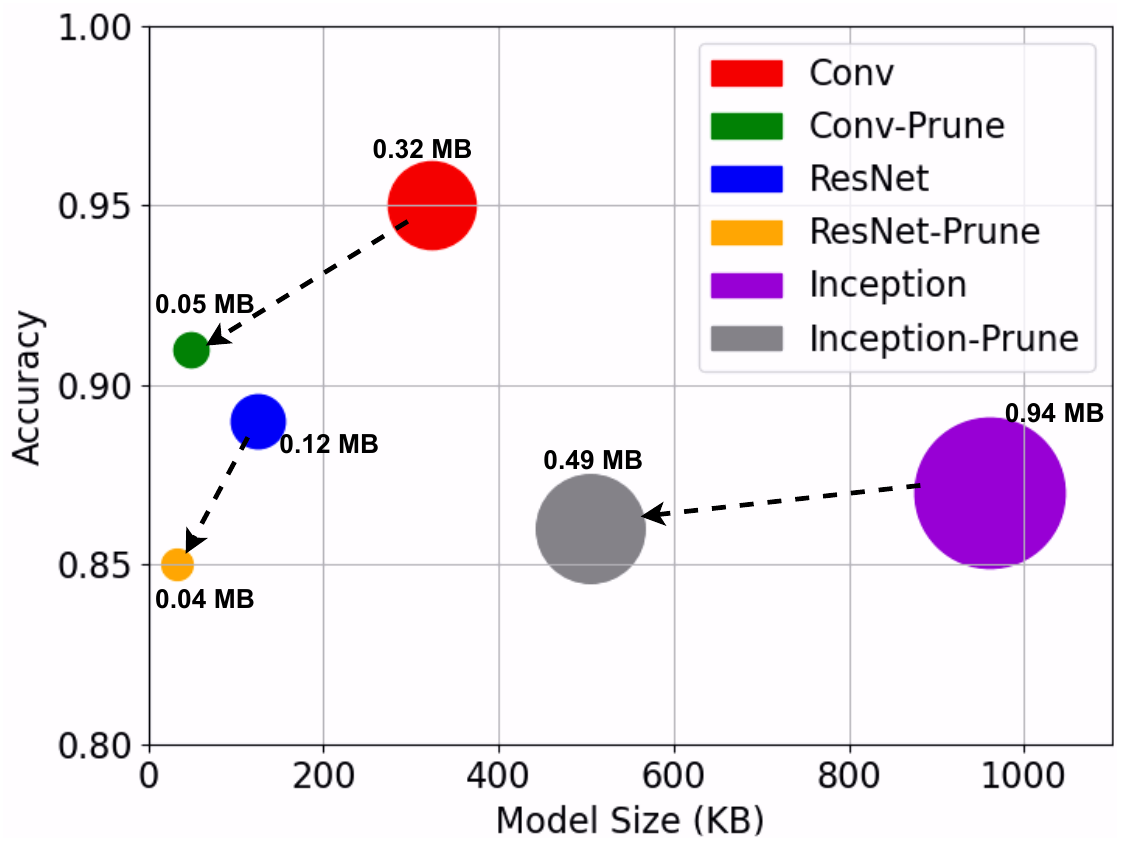}
    \caption{The diagram illustrates the model size and accuracy for various models. The size of each circle and the value adjacent to it represent the memory consumption (RAM) required for running the inference. Directed arrows show the transition from the original model to its pruned counterpart.}
    \label{fig:memory_usage}
\end{figure}

\begin{figure*}[!ht]
    \centering
    \includegraphics[width=0.8\textwidth]{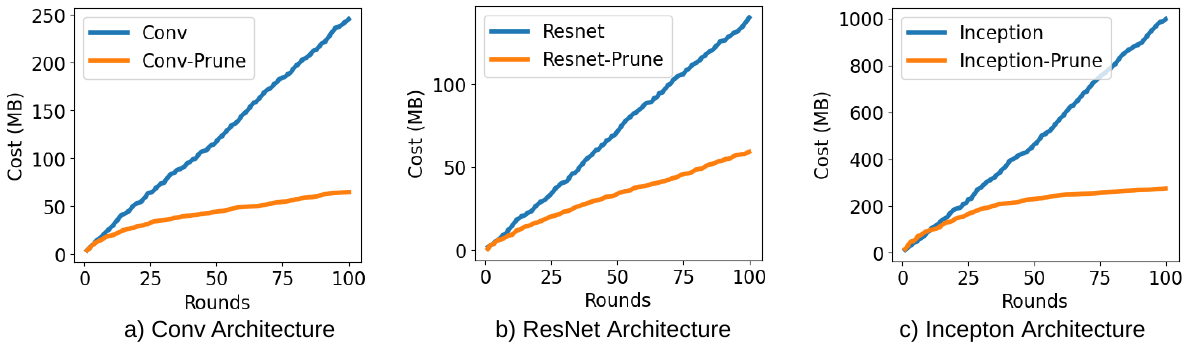}
    \caption{Communication Cost across different convolutional architectures on the FEMNIST dataset.}
    \label{fig:communication_cost}
\end{figure*}

\begin{figure*}[ht]
    \centering
    \includegraphics[width=0.8\textwidth]{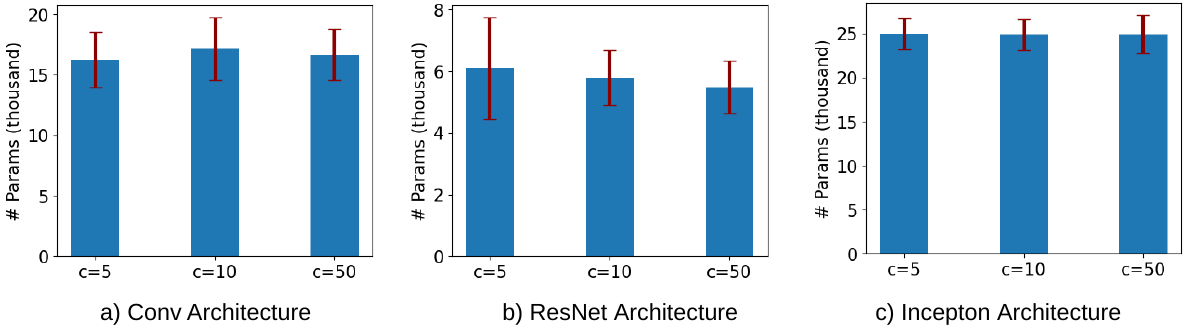}
    \caption{Uncertainty in model pruning across different client configurations on the FL system (c is the number of selected clients). Each error bar represent the mean number of parameters resulting from 10 different experimental runs.}
    \label{fig:client_drop}
\end{figure*}

When deploying deep learning models into real-world applications, it is crucial to consider several factors beyond inference time. Figure \ref{fig:memory_usage} illustrates a comparison between the original and pruned models based on several criteria, including model size, accuracy, and memory usage.  The experiments were conducted using the tflite model format on an Ubuntu Dell Desktop. The outcomes demonstrate that in all tested architectures, the proposed pruning method significantly reduced the required disk space when compared to their original counterpart, making our solution more suitable for devices with storage constraints.  Additionally, memory consumption (RAM) during the inference phase was significantly lower in the pruned models. This reduction in both model size and memory usage enhances the feasibility of deploying these models on devices with limited resources.

\subsection{Communication Cost and Client Drop}

In this section, we evaluate communication cost, which is defined as the total amount of data transferred between server and clients during the FL training process.
For this experiment, we conducted 100 rounds of FL training and calculated the total communication cost by summing the data transferred in all rounds.
The data transferred is determined by the size of the model when sent from the server to the client and vice versa.
Specifically, the model size is calculated by multiplying the number of model parameters by the size of each parameter element. 
In our default setup, we use float32 elements, which are 4 bytes each.

Figure \ref{fig:communication_cost} illustrates the communication cost across various convolutional architectures, including Convolution, ResNet, and Inception on the FEMNIST dataset. 
In all cases, our pruning method significantly reduces the communication cost compared to normal training.
Notably, in the experiments involving vanilla Conv (Figure~\ref{fig:communication_cost}.a) and Inception (Figure~\ref{fig:communication_cost}.c), the cost is reduced up to 5 times. 
This reduction is a result of our model being progressively  pruned during the FL training process.
Consequently, the required communication cost diminishes and enhances the overall efficiency of the FL process.

In the FL systems, the number of clients participating in each round can vary due to various factors like availability and connectivity. 
To ensure the robustness of our proposed pruning technique, we conducted a series of experiments. 
Specifically, we tested the pruning technique with different numbers of selected clients, including 5, 10, and 50 clients per round, which allowed us to evaluate how well the pruning technique performs under different number of client participation.
These experiments were performed using the FEMNIST dataset and involved various neural network architectures (Conv, ResNet, and Inception architecture).

The error bar in Figure~\ref{fig:client_drop} visualizes the uncertainty associated with the performance of a pruning algorithm in FL system across different numbers of clients.
The x-axis represents the different client numbers per round in the FL training process and the y-axis represents the number parameters of the pruned model.
On each architecture, the mean value of the retained parameters (represented by the height of each bar) remains approximately the same value when training with different numbers of selected clients.
This consistency indicates that our pruning method maintains stable performance regardless of the number of clients participating in each training round. 
This robustness is particularly noteworthy given the inherent variability in the Federated Learning (FL) process.

% The box plots in Figure~\ref{fig:client_drop} provide a visualization regarding the distribution of model sizes.
% The x-axis represents the different client numbers per round in the FL training process and the y-axis represents the model size of the pruned model.
% On each architecture, the median (represented by the red line) remains approximately the same value when training with different numbers of selected clients.
% This consistency in the median value indicates that our pruning method maintains stable performance regardless of the number of clients participating in each training round. 
% This robustness is particularly noteworthy given the inherent variability in the Federated Learning (FL) process.

\subsection{Ablation Study Hyper-parameter $k$}

\begin{figure}[h]
    \centering
    \includegraphics[width=0.35\textwidth]{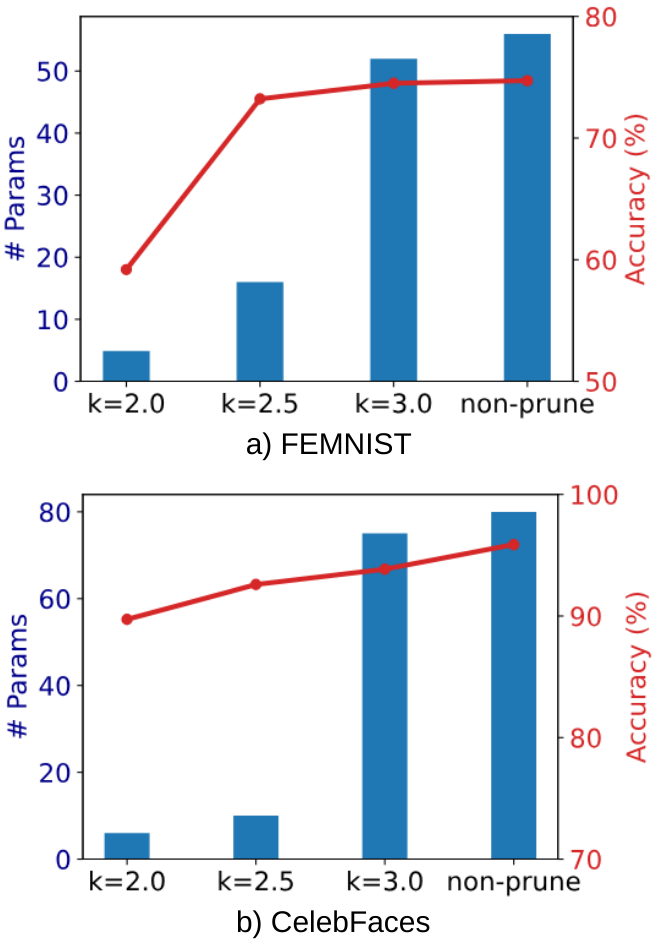}
    \caption{The Pareto Diagram shows the trade-off between pruning aggressive and accuracy retention across different values of hyper-parameter $k$.}
    \label{fig:ablation}
\end{figure}

In this subsection, we conduct an ablation study to deepen our understanding of how the hyper-parameter $k$ affects the performance of pruning Algorithm~\ref{algo:prune}.
Specifically, we perform experiments using the convolutional architecture on two datasets FEMNIST and CelebFaces with $k$ values set at 2.0, 2.5, and 3.0 respectively.
The Pareto Diagram, shown in Figure~\ref{fig:ablation}, illustrates the trade-off between a number of parameters (in thousands) and model accuracy, demonstrating the impact of varying $k$.

The bar chart, colored in blue and plotted on the left y-axis, represents the number of parameters of the pruned model, illustrating the aggressiveness of the pruning process at each value of $k$. As $k$ decreases, a greater proportion of the model is pruned, reflecting a more aggressive reduction in model size.
Interestingly, when $k=3$, the pruned model retains nearly the same number of parameters as the original (non-prune) model.
Conversely, the line chart on the right y-axis (color in red) displays the corresponding model accuracy. As expected, a clear trend emerges where higher pruning typically corresponds with lower accuracy levels. 
This pattern highlights the essential trade-offs that need to be considered when optimizing the hyper-parameter $k$ for pruning.

\section{Conclusion}
\label{sec_conclusion}

In this paper, we presented a structured pruning method for Federated Learning (FL) systems that significantly reduces the model's parameters, computational costs, and communication overhead without introducing sparsity into the weight matrix.
In addition, to tackle the challenge of choosing an optimal model architecture in FL environments, we introduced an automatic pruning algorithm that determines the optimal number of filters to prune.
Our method effectively reduces up to 90\% of parameters and FLOPS on the FEMNIST dataset and 80\% on the CelebFaces dataset, with only a minimal loss in accuracy.
Additionally, when deployed on real Android devices, the pruned model cuts inference time by up to 50\% and doubles the throughput.
Furthermore, the pruned model decreases communication costs up to five times and maintains consistent performance across varying numbers of selected clients during the FL procedure.

\setcounter{secnumdepth}{1}
\appendix

\setcounter{secnumdepth}{0}

\nobibliography{aaai22}
\section{Reference}
\bibentry{babakniya2023revisiting}.\\[.2em] 
\bibentry{Bibikar_Vikalo_Wang_Chen_2022}.\\[.2em]
\bibentry{caldas2018leaf} .\\[.2em]
\bibentry{dai2022dispfl} .\\[.2em]
\bibentry{glorot2010understanding} .\\[.2em]
\bibentry{han2015learning} .\\[.2em]
\bibentry{he2015delving} .\\[.2em]
\bibentry{he2016deep} .\\[.2em]
\bibentry{huang_distributed_2023} .\\[.2em]
\bibentry{hyeon2021fedpara} .\\[.2em]
\bibentry{isik2023sparse} .\\[.2em]
\bibentry{jiang_model_2023} .\\[.2em]
\bibentry{jiang2023computation} .\\[.2em]
\bibentry{jiang2023complement} .\\[.2em]
\bibentry{jia2024fedlps}.\\[.2em]\
\bibentry{jin2024fractional} .\\[.2em]
\bibentry{li2017pruning} .\\[.2em]
\bibentry{li2021lotteryfl} .\\[.2em]
\bibentry{li2021fedmask} .\\[.2em]
\bibentry{li2021talk}.\\[.2em]\
\bibentry{liu2018rethinking} .\\[.2em]
\bibentry{mcmahan2017communication}.\\[.2em]\
\bibentry{qiu2022zerofl} .\\[.2em]
\bibentry{slim_2017} .\\[.2em]
\bibentry{szegedy2015going} .\\[.2em]
\bibentry{vahidian2021personalized} .\\[.2em]
\bibentry{wu2020fedscr} .\\[.2em]
\bibentry{xu2021accelerating} .\\[.2em]
\bibentry{zhang2022fedduap}.\\[.2em]\

\end{document}